\documentclass[letterpaper]{article}
\usepackage{aaai19}
\usepackage{times}
\usepackage{helvet}
\usepackage{courier}
\usepackage[hyphens]{url}
\usepackage{graphicx}
\usepackage{graphicx}
\usepackage{pifont}
\usepackage{xcolor}
\usepackage{xspace}

\newcommand{\eg}{\emph{e.g.,}\xspace}

\begin{document}

\title{Component Mismatches Are a Critical Bottleneck to Fielding AI-Enabled Systems in the Public Sector}
\author{Grace A. Lewis, Stephany Bellomo, April Galyardt\\
\{glewis, sbellomo, akgalyardt\}@sei.cmu.edu\\
Carnegie Mellon Software Engineering Institute\\
Pittsburgh, PA USA
}
\maketitle
\begin{abstract}
The use of machine learning or artificial intelligence (ML/AI) holds substantial potential toward improving many functions and needs of the public sector. In practice however, integrating ML/AI components into public sector applications is severely limited not only by the fragility of these components and their algorithms, but also because of mismatches between components of ML-enabled systems. For example, if an ML model is trained on data that is different from data in the operational environment, field performance of the ML component will be dramatically reduced. Separate from software engineering considerations, the expertise needed to field an ML/AI component within a system frequently comes from outside software engineering. As a result, assumptions and even descriptive language used by practitioners from these different disciplines can exacerbate other challenges to integrating ML/AI components into larger systems. We are investigating classes of mismatches in ML/AI systems integration, to identify the implicit assumptions made by practitioners in different fields (data scientists, software engineers, operations staff) and find ways to communicate the appropriate information explicitly. We will discuss a few categories of mismatch, and provide examples from each class. To enable ML/AI components to be fielded in a meaningful way, we will need to understand the mismatches that exist and develop practices to mitigate the impacts of these mismatches.
\end{abstract}

\section{Introduction}
The public sector owns and uses many very large data collections for a variety of purposes. Machine learning or artificial intelligence (ML/AI) components hold substantial potential toward improving many functions and needs of the public sector, based on the analysis of this data. In practice however, fielding these components into public sector applications is severely limited by the fragility of ML/AI components and their algorithms. For systems that touch the government and public sector, some considerations for ML/AI systems are more at the forefront than in the commercial sector, including privacy, security and ethics. 

One of the challenges in deploying complex systems is integrating all components and resolving any component mismatches. For systems that incorporate ML/AI components, the sources and effects of these mismatches may be different from other software integration efforts. For example, if an ML model is trained on data that is different from data in the operational environment, field performance of the ML component will be dramatically reduced. Separate from software engineering considerations, the expertise needed to field an ML/AI component within a system frequently comes from outside software engineering. As a result, assumptions and even descriptive language used by practitioners from these different disciplines can exacerbate other challenges to integrating ML/AI components into larger systems. 

We are investigating classes of mismatches in ML/AI systems integration, to identify the implicit assumptions made by practitioners in different fields (data scientists, software engineers, operations staff) and find ways to communicate the appropriate information explicitly. We will discuss a few categories of mismatch, and provide examples from each class. To enable ML/AI components to be fielded in a meaningful way, we will need to understand the mismatches that exist and develop practices to mitigate the impacts of these mismatches. This paper reports on the goals of our study and the expected results.

\section{Mismatch in Machine-Learning Enabled Systems}

Despite the growing interest in ML and AI across all industries --- including DoD, government, and public sector --- development of ML and AI capabilities is still mainly a research activity or a stand-alone project, with the exception of large companies such as Google and Microsoft \cite{Ghelani2019}. Deploying ML models in operational systems remains a significant challenge \cite{Amershi2019}\cite{Sculley2015}\cite{Talby2018}. 

\begin{figure}[htbp]
\centerline{\includegraphics[width=\linewidth]{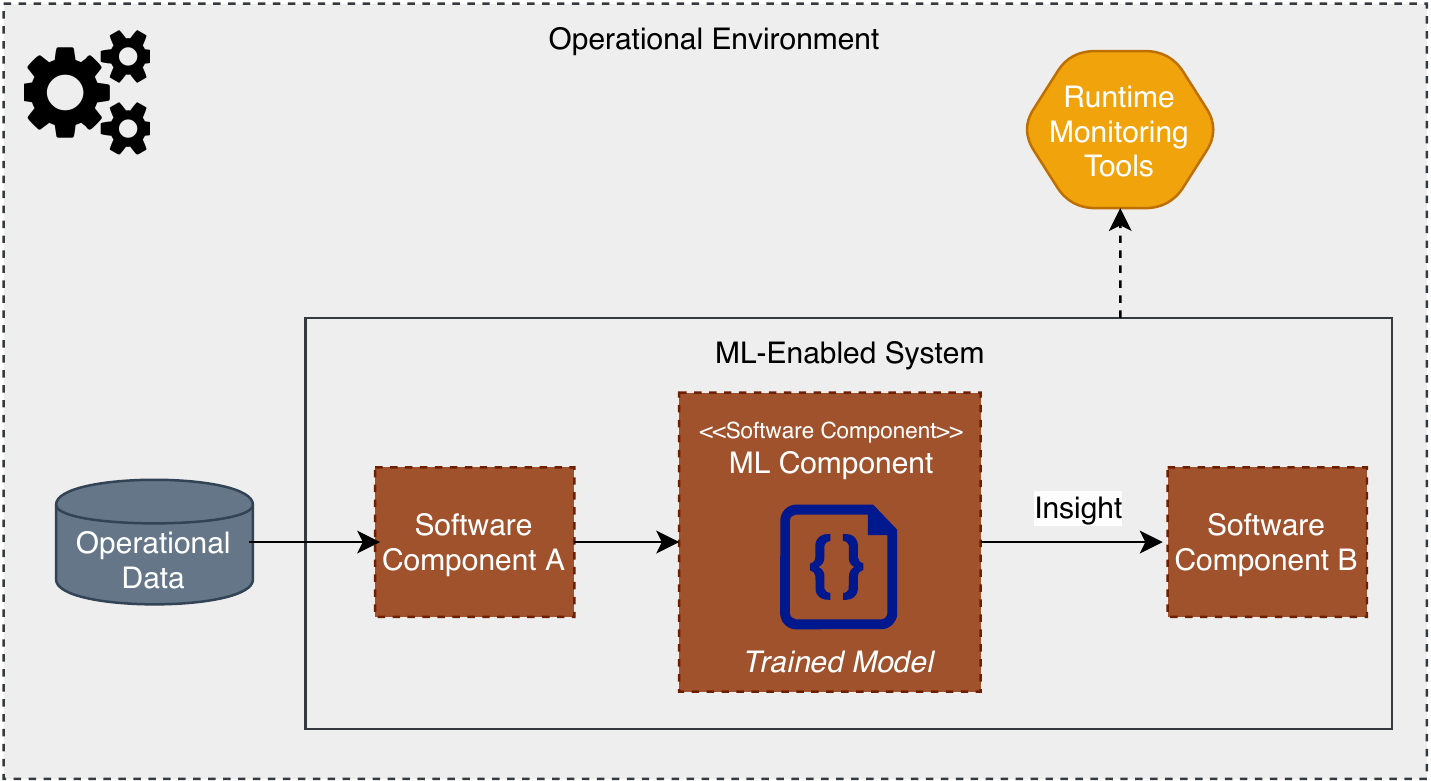}}
\caption{Elements of a Deployed ML-Enabled System and Operations Perspective}
\label{fig-ml-enabled-system}
\end{figure}

We define an ML-enabled system as a software system that relies on one or more ML software components to provide required capabilities, as shown in Figure  \ref{fig-ml-enabled-system}. The ML component in this figure receives (processed) operational data from one software component and generates an insight that is consumed by another software component. A problem in these types of systems is that their development and operation involve three perspectives, with three different and often completely separate workflows and people.

\begin{figure}[htbp]
\centerline{\includegraphics[width=\linewidth]{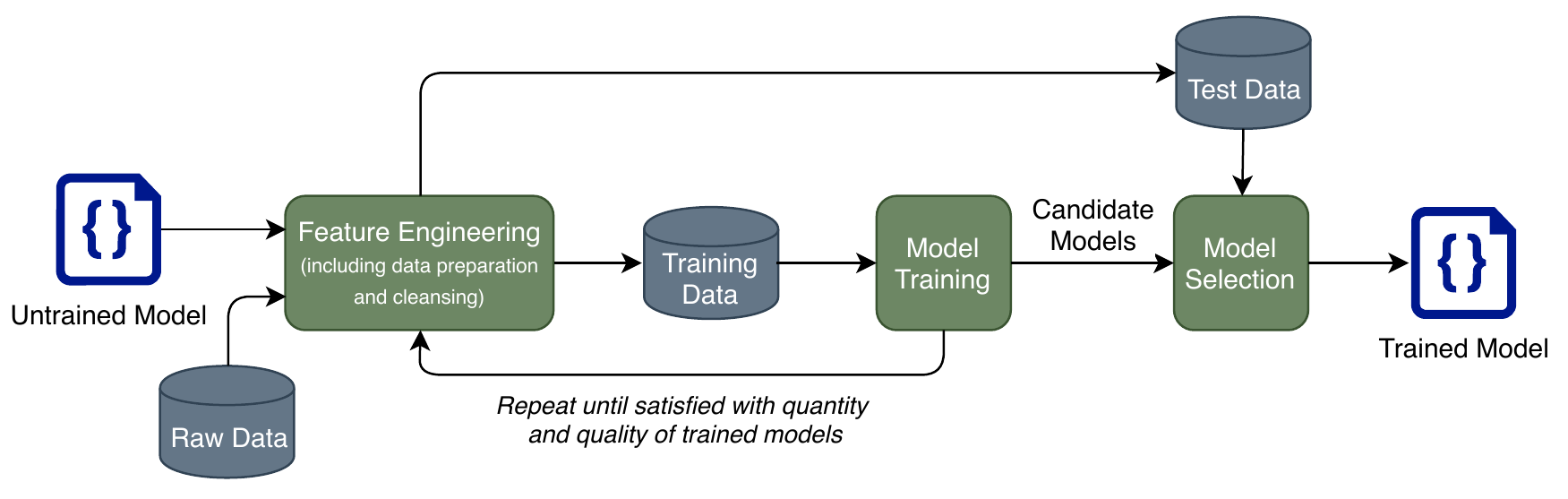}}
\caption{Data Scientist Perspective}
\label{fig-ml-data-scientist}
\end{figure}

\begin{figure}[htbp]
\centerline{\includegraphics[width=\linewidth]{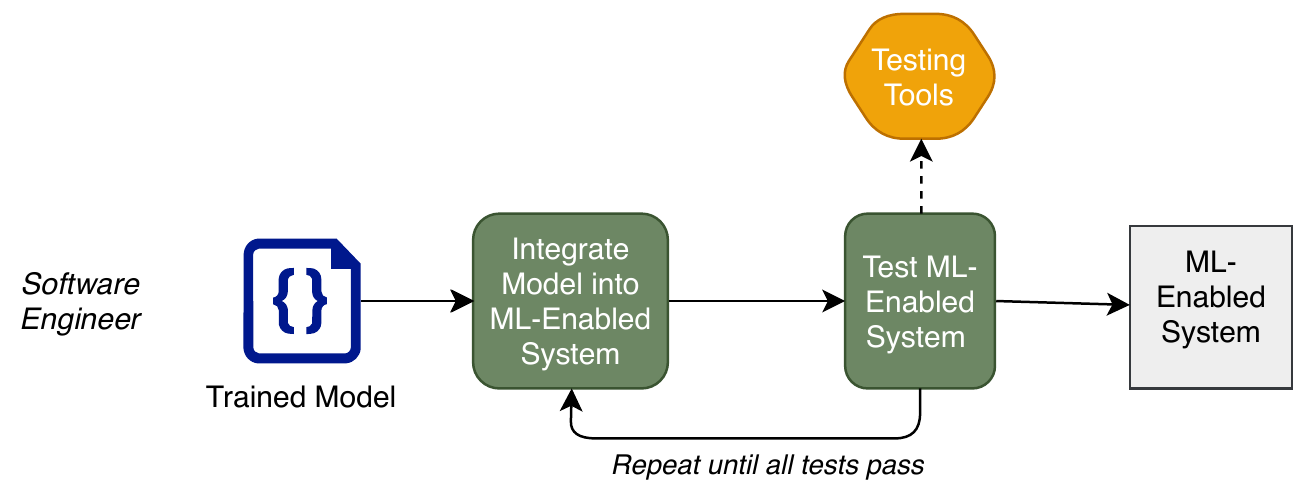}}
\caption{Software Engineer Perspective}
\label{fig-ml-software-engineer}
\end{figure}

\begin{enumerate}
\item The \textit{Data Scientist} builds the model: The workflow of the data scientist, as shown in Figure \ref{fig-ml-data-scientist}, is to take an untrained model and raw data, use feature engineering to create a set of training data that is then used to train the model, repeating these steps until a set of adequate models are produced, and then using a set of test data to test the different models and select the one that performs the best based on a set of defined evaluation metrics. Out of this workflow comes a trained model. 
\item The \textit{Software Engineer} integrates the trained model into a larger system: The workflow of the software engineer, as shown in Figure \ref{fig-ml-software-engineer}, is to take the trained model, integrate the model into the ML-enabled system, and test the system until it passes all tests. The ML-enabled system is then passed to operations staff for deployment.
\item \textit{Operations Staff} deploy, operate, and monitor the system: As shown in Figure \ref{fig-ml-enabled-system}, in addition to the operation and monitoring of the ML-enabled system, operations staff are also responsible for operation and monitoring of operational data sources (\eg databases, data streams, data feeds).
\end{enumerate}

Because these perspectives operate separately and often speak different languages, there are opportunities for mismatch between the assumptions made by each perspective with respect to the elements of the ML-enabled system, and the actual guarantees provided by each element. This problem is exacerbated by the fact that system elements evolve independently and at a different rhythm, which could over time lead to unintentional mismatch. In addition, we expect these perspectives to belong to three different organizations, especially in the public sector. Examples of mismatch and their consequences include

\begin{itemize}
    \item poor system performance because computing resources used during testing of the model are different from computing resources used during operations
    \item poor model accuracy because model training data is different from operational data
    \item development of large amounts of glue code because the trained model input/output is incompatible with operational data types
    \item system failure due to inadequate testing because developers were not able to replicate the testing that was done during model training
    \item monitoring tools are not set up to detect diminishing model accuracy, which is the “performance” metric defined for the trained model.
\end{itemize}

\section{ML-Enabled System Element Descriptors}

We are developing machine-readable \textit{ML-Enabled System Element Descriptors} as a mechanism to enable mismatch detection and prevention in ML-enabled systems. The goal of the descriptors is to codify attributes of system elements and therefore make explicit all assumptions from all perspectives. The descriptors can be used by system stakeholders in a manual way, for information, awareness and evaluation activities; and by automated mismatch detectors at design time and runtime for cases in which attributes lend themselves to automation.  While there is existing, recent work in creating descriptors for data sets \cite{Gebru2018}, models \cite{Mitchell2019}, and online AI services \cite{Hind2018}, there are two main limitations in this work: (1) they do not address the software engineer and operations perspectives, and (2) they are not machine-readable. Our work addresses these two limitations, in addition to providing the following immediate benefits:
\begin{itemize}
\item Definitions of mismatch can serve as checklists as ML-enabled systems are developed
\item Recommended descriptors provide stakeholders (\eg program offices) with examples of information to request and/or requirements to impose
\item Means identified for validating ML-enabled system element attributes provide ideas for confirming information provided by third-parties
\item Identification of attributes for which automated detection is feasible defines new software components for ML-enabled systems 
\end{itemize}

\section{Study Protocol}
The technical approach for constructing and validating the ML-Enabled System Element Descriptors consists of three phases.

\begin{figure}[htbp]
\centerline{\includegraphics[width=\linewidth]{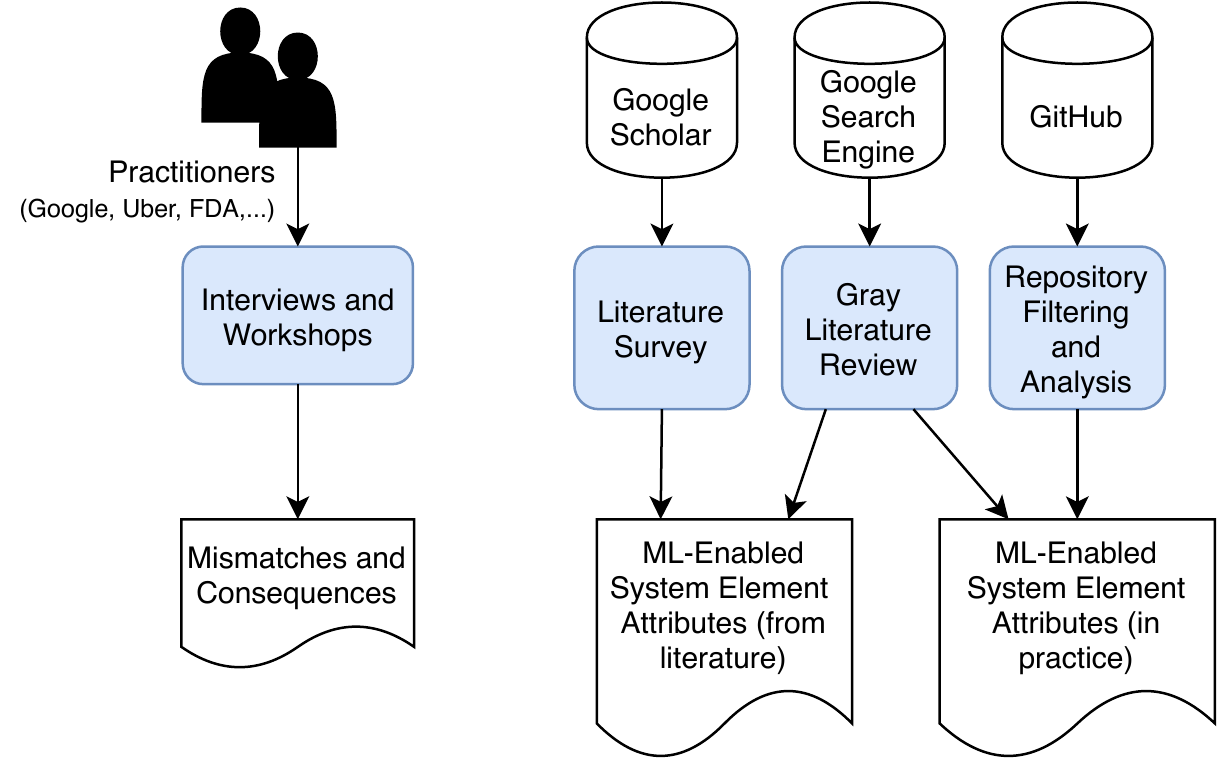}}
\caption{Information Gathering}
\label{fig-info-gathering}
\end{figure}

\textbf{Phase 1 - Information Gathering:}  As shown in Figure \ref{fig-info-gathering}, this phase involves two parallel tasks. In one task, we elicit examples of mismatches and their consequences from practitioners via interviews and/or workshops. In the second task, we identify attributes currently used to describe elements of ML-enabled systems by mining project descriptions from GitHub repositories that contain trained and untrained models \cite{Kalliamvakou2016}, a literature survey, and a gray literature review \cite{Garousi2019}. This multi-modal approach provides both the practitioner and the academic perspective on best practices to describe ML-enabled system elements. 

\begin{figure}[htbp]
\centerline{\includegraphics[width=2in]{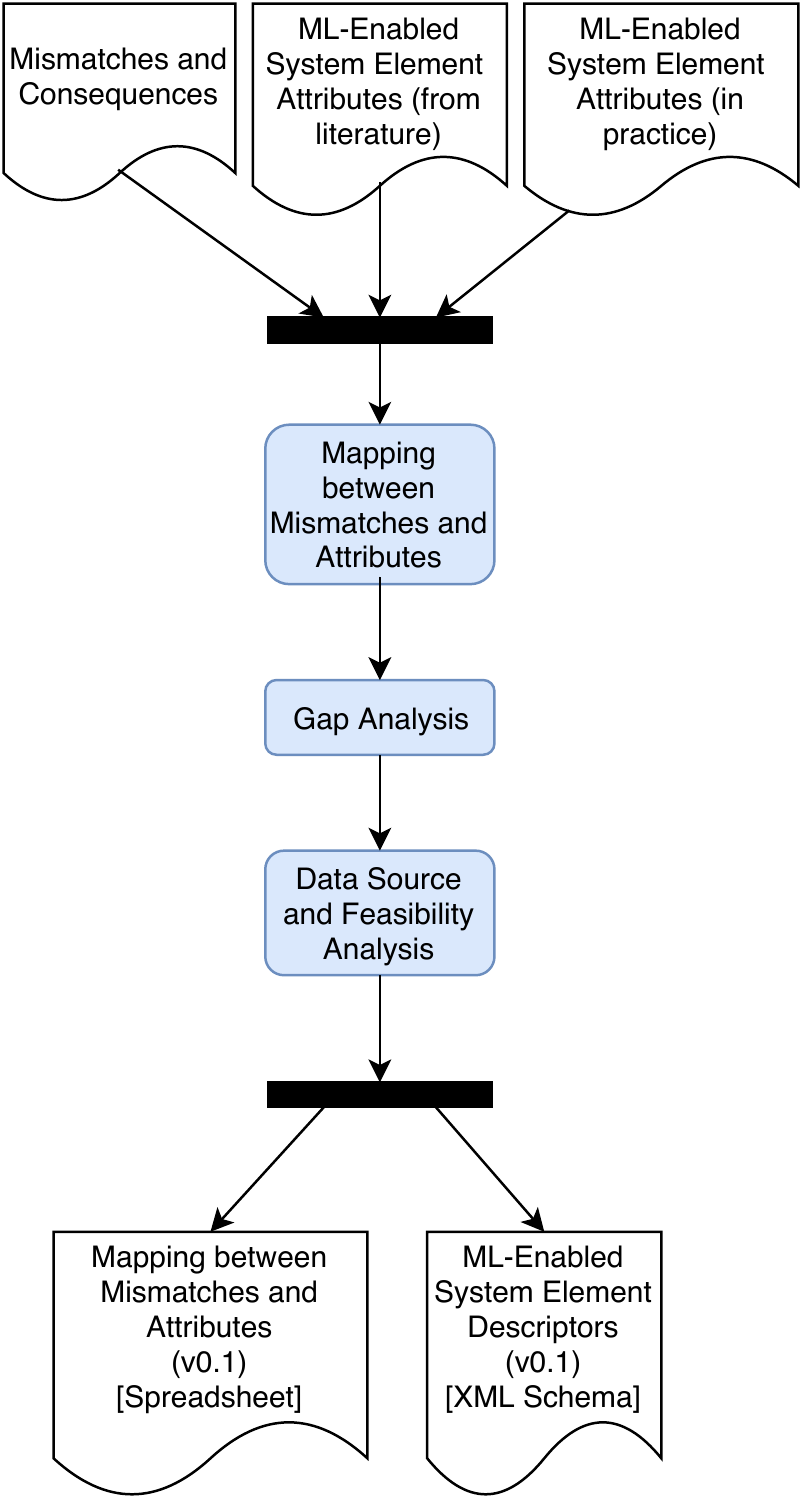}}
\caption{Analysis}
\label{fig-analysis}
\end{figure}

\begin{figure*}[t]
\centerline{\includegraphics[width=\linewidth]{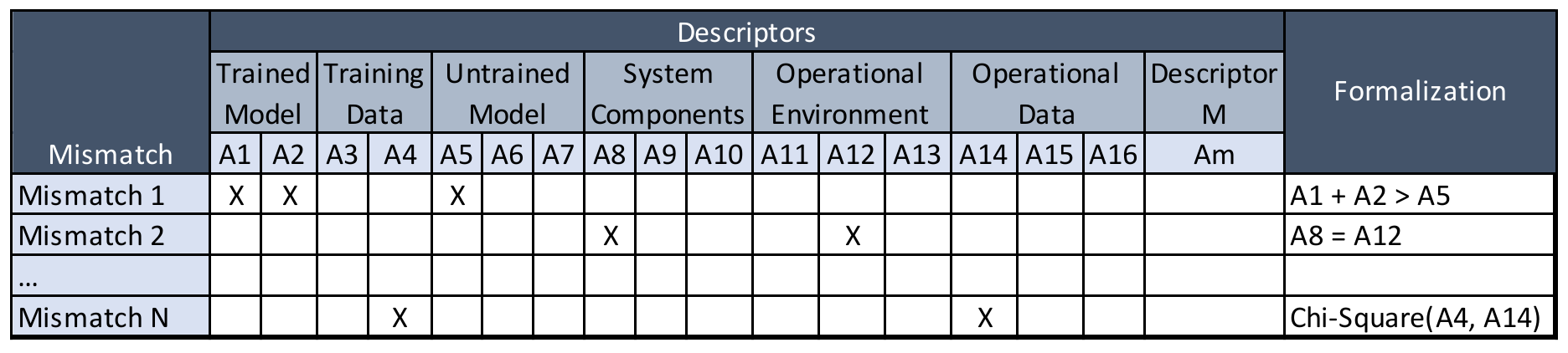}}
\caption{Mapping between Mismatches and ML-Enabled System Element Attributes}
\label{fig-matrix}

\end{figure*}

\textbf{Phase 2 - Analysis:} The tasks in this phase are shown in Figure \ref{fig-analysis}. Once mismatches and attributes of elements of ML-enabled systems have been elicited, there is a mapping stage in which an initial version of the spreadsheet shown in Figure \ref{fig-matrix} is produced. For each mismatch we identify the set of attributes that could be used to detect that mismatch, and formalize the mismatch as a predicate over those attributes, as shown in the Formalization column in Figure \ref{fig-matrix}. As an example, the figure shows that Mismatch 1 occurs when the value of Attribute 1 plus the value of Attribute 2 is greater than the value of Attribute 5. The second step is to perform gap analysis to identify mismatches that do not map to any attributes and attributes that do not map to any mismatch. We then complement the mapping based on our domain knowledge, by adding attributes and potentially adding new mismatches that could be detected based on the available attributes. Finally, there is a data source and feasibility analysis step where for each attribute we identify the data source (who provides the value), the feasibility of collecting those values  (is it reasonable to expect someone to provide that value and/or is there a way of automating its collection), how can it be validated (if necessary to validate that the provided value is correct), and finally potential for automation (can the set of identified attributes be used in scripts or tools for detecting that mismatch). After the analysis stage we have an initial version of the spreadsheet, and of the descriptors derived from the spreadsheet.

\begin{figure}[htbp]
\centerline{\includegraphics[width=\linewidth]{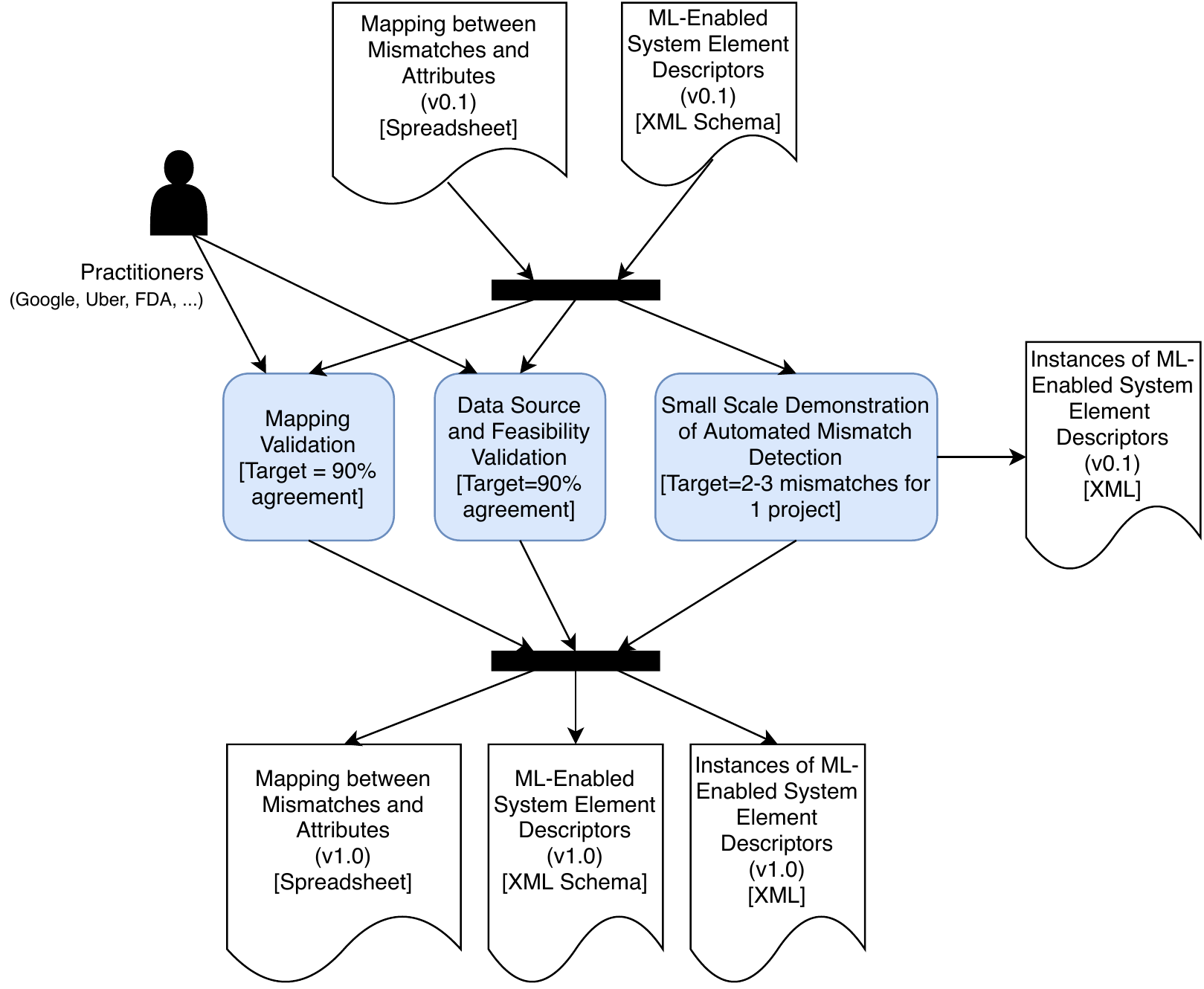}}
\caption{Evaluation}
\label{fig-evaluation}
\end{figure}

\textbf{Phase 3 – Evaluation:} As shown in Figure \ref{fig-evaluation}, in this stage we re-engage with the interview or workshop participants from the Information Gathering stage to validate mapping, data sources, and feasibility. The evaluation target is a 90\% agreement on the work developed in the Analysis stage. In addition, we develop a small-scale demonstration of automated mismatch detection. The target is to identify 2-3 mismatches in a project that can be detected via automation, and develop scripts that can detect the mismatch. In the end, the project outcomes are the validated mapping between mismatches and attributes, a set of descriptors created from that mapping, and instances of the descriptors.

\section{Summary and Next Steps}

Our vision for this work is that the community starts developing tools for automatically detecting mismatch, and organizations start including mismatch detection in their toolchains for development of ML-enabled systems. As a step toward this vision, we are working on the following artifacts:
\begin{itemize}
\item List of mismatches in ML-enabled systems and their consequences
\item List of attributes for ML-enabled system elements 
\item Mapping of mismatches to attributes (spreadsheet)
\item XML schema for each descriptor (one per system element) plus XML examples of descriptors
\item Small-scale demonstration (scripts) of automated mismatch detection
\end{itemize}

We are using opportunities such as this workshop to (1) socialize the concept of mismatch and convey its importance for the deployment of ML-enabled systems into production, (2) elicit and confirm mismatches in ML-enabled systems and their consequences from people in the field, in particular from the public sector, (3) obtain early feedback on the study protocol and resulting artifacts.

\section*{Acknowledgements}

This material is based upon work funded and supported by the Department of Defense under Contract No. FA8702-15-D-0002 with Carnegie Mellon University for the operation of the Software Engineering Institute, a federally funded research and development center (DM19-1007).

\bibliographystyle{aaai}
\bibliography{ml-mismatch}

\end{document}